\def\year{2018}\relax
\documentclass[letterpaper]{article} 
\usepackage{aaai18}  
\usepackage{times}  
\usepackage{helvet}  
\usepackage{courier}  
\usepackage{url}  
\usepackage{graphicx}  
\usepackage{amssymb}
\usepackage{mathtools}
\usepackage{color}
\usepackage{algorithm}
\usepackage{algorithmic}
\begin{document}
	%
	\title{Curve-Structure Segmentation from Depth Maps: A CNN-based Approach and Its Application to Exploring Cultural Heritage Objects}
	\author{Yuhang Lu\textsuperscript{$2$}, Jun Zhou\textsuperscript{$2$}, Jing Wang\textsuperscript{$2$}, Jun Chen\textsuperscript{$2$}, Karen Smith\textsuperscript{$3$}, Colin Wilder\textsuperscript{$4$}, Song Wang\textsuperscript{$1,2,\thanks{Corresponding author: songwang@cec.sc.edu}$}\\
		\textsuperscript{$1$} School of Computer Science and Technology, Tianjin University, Tianjin, China\\
		\textsuperscript{$2$} Department of Computer Science and Engineering, University of South Carolina, Columbia, SC\\
		\textsuperscript{$3$} South Carolina Institute of Archaeology and Anthropology, University of South Carolina, Columbia, SC\\
		\textsuperscript{$4$} Center for Digital Humanities, University of South Carolina, Columbia, SC\\}
	\maketitle
	\begin{abstract}
	Motivated by the important archaeological application of exploring cultural heritage objects, in this paper we study the challenging problem of automatically segmenting curve structures that are very weakly stamped or carved on an object surface in the form of a highly noisy depth map. Different from most classical low-level image segmentation methods that are known to be very sensitive to the noise and occlusions, we propose a new supervised learning algorithm based on Convolutional Neural Network (CNN) to implicitly learn and utilize more curve geometry and pattern information for addressing this challenging problem. More specifically,  we first propose a Fully Convolutional Network (FCN) to estimate the skeleton of curve structures and at each skeleton pixel, a scale value is estimated to reflect the local curve width. Then we propose a dense prediction network to refine the estimated curve skeletons. Based on the estimated scale values, we finally develop an adaptive thresholding algorithm to achieve the final segmentation of curve structures. In the experiment, we validate the performance of the proposed method on a dataset of depth images scanned from unearthed pottery sherds dating to the Woodland period of Southeastern North America.
	\end{abstract}
	
	\section{Introduction}
	
\begin{figure}[htbp]
	\centering
	\includegraphics[width=0.48\textwidth]{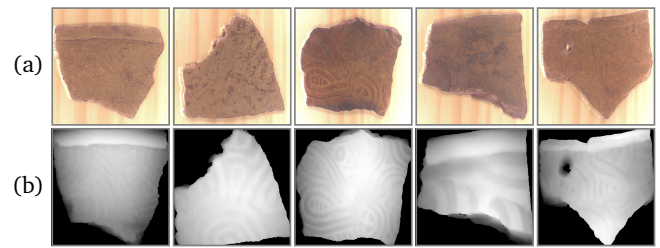}
	\caption{Five unearthed pottery sherds dating to the Woodland period of Southeastern North America. (a) RGB images. (b) Depth images where intensity indicates the depth.}	\label{fig1}
	\end{figure}

	Embellished designs on the surface of cultural heritage objects, such as pottery, shell, stone and wood contain important information for archaeologists \cite{zhou2017identifying}. These designs, if successfully identified and correlated, can be used to build chronologies and track trade networks of a region thousands of years ago. In archeology, most of these designs are found to be curve patterns stamped or carved by their makers. Therefore, it is of great interest to archaeologists to accurately segment the curve structures on the surface of unearthed fragments of cultural heritage objects and identify their underlying designs~\cite{kampel2007rule,halir1999automatic}. Figure~\ref{fig1} shows several unearthed pottery sherds dating to the Woodland period of Southeastern North America. The curve structures on their surfaces reflect a portion of the curve pattern carved into wooden paddles and applied onto hand-built clay vessels designed by southeastern Native Americans around 2000 years ago.  There are hundreds of thousands of such fragmented culture heritage objects stored in museums, which calls for more intelligent and automatic tools to explore them.
	
	Clearly, accurately segmenting the curve structures stamped on the surface is the first step to explore these cultural heritage objects. In most cases, these curve structures do not bear distinctive colors and it is very difficult, if not impossible, to segment them from an RGB image of the sherd, e.g., Figure~\ref{fig1}(a), taken by traditional cameras. In archeology, 3D scanners are usually utilized to produce a depth image of the object surface -- with paddle stamping, the locations of curves exhibit a larger depth than the non-curve portion of surface, as shown in Figure~\ref{fig1}(b). 
	
\begin{figure}[htbp]
	\centering
	\includegraphics[scale = 0.4]{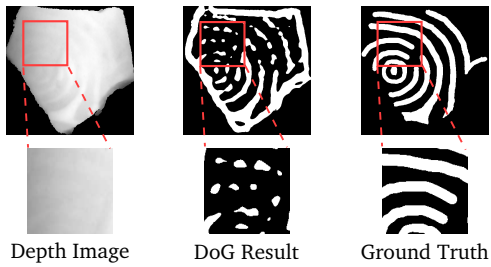}
	\caption{An illustration of using low-level methods for curve-structure segmentation. Serious erosion in the red square leads to very low contrast in the depth image, and low-level method, such as DoG (Difference of Gaussian), may produce very poor segmentation.}\label{fig2}
	\end{figure}

	However, three complexities may lead to very weak curve structures on the obtained depth map and make the curve-structure segmentation a very challenging problem. First, the carved paddles used for stamping are usually flat while the object surfaces are usually not. As a result, the paddle typically does not well fit the object surface, which leads to shallow curves at many locations. Second, purposeful smoothing of the stamped surface during vessel manufacture or weathering and erosion after vessel discard can lead to subtle depth differences between the curve and the non-curve portions of the surface. Third, erosion and weathering make the object surface highly rough, which is equivalent to adding random noise to the depth map of the initial object surface. With these three complexities, it is difficult to use a low-level image segmentation algorithm to accurately segment these depth images for curve structures, as shown by an example in Figure~\ref{fig2}.

	In this paper, we propose a new supervised learning approach to segment such curve structures that were weakly stamped on object surface.  The basic idea is that, in most applications, such as exploring cultural heritage objects in archeology, the underlying designs of the curve structures bear certain geometries and patterns. For example, most of the curve structures consist of smooth curve segments. Furthermore, many curves in the structures show good parallelism against each other. These characteristics give the material a visually distinctive style \cite{smith2012style}. Consideration of these high-level geometry and pattern information may help improve the accuracy and reliability of curve-structure segmentation. While it is difficult to handcraft the features of all relevant curve geometry and pattern in an application, we expect the proposed approach can automatically learn these features from a set of training data with labeled ground truth.

    In practice, the curve structures of interest have width, which may vary along the curve and need to be inferred in segmentation. However, it is well known that the curve geometry and pattern are independent of the curve width. Mixing all of them may substantially increase the difficulty of feature learning for segmentation. In this paper, we handle them separately by developing a three-step curve-structure segmentation algorithm. In the first step, a Fully Convolutional Network (FCN) is employed to extract the skeleton of curve structures, and estimate a scale value at each skeleton pixel. This scale value reflects the curve width at the corresponding skeleton pixel. In the second step, we propose a dense prediction network to refine the curve skeletons. In the third step, we develop an adaptive thresholding algorithm to achieve the final segmentation of curve structures with width by considering the estimated scale values.

    For the experiments, we collected the depth image of a set of pottery sherds excavated from archaeological sites associated with the Swift Creek paddle-stamped tradition of southeastern North America. Ground-truth curve structure segmentation are manually constructed. We evaluate the proposed method on the collected depth images and compare its performance against several other existing algorithms. We also evaluate the segmentation results in the task of design matching in archeology.
	
	\section{Related Work}
	
	\textbf{General-purpose image segmentation} has been studied for many decades, resulting in many image segmentation algorithms. For example, by considering only low-level pixel intensities, many edge detection \cite{wang2004salient,arbelaez2011contour}, region growing/splitting \cite{tremeau1997region}, pixel clustering \cite{li2015superpixel}, and graph-based algorithms \cite{shi2000normalized,wang2001image,wang2003image} have been developed for segmenting an image into multiple regions. By further considering mid-level cues like boundary smoothness, many active-contour and level-set algorithms have been developed to segment foreground objects from background \cite{chan2001active,vese2002multiphase}. In principle, these general-purpose image segmentation algorithms can be easily adapted to handle our problem of segmenting curve structures from depth images, by treating depth value as intensity. However, their segmentation performances are usually poor when the depth image is noisy and the desired curve structures are weak. In the experiments, we include several general-purpose segmentation algorithms, such as DoG, LevelSet \cite{vese2002multiphase}, and GrabCut \cite{rother2004grabcut}, as comparison methods.
	
	\textbf{Deep-learning based algorithms}, particularly the CNN-based algorithms, have been recently used for image segmentation, by learning high-level features of the desired segments in a supervised way \cite{segnet,crfasrnn}. The most influential one is the Fully Convolutional Network proposed by~\citeauthor{fcn}~\shortcite{fcn}. It transforms traditional fully connected layers to convolution layers, thus enabling to train and predict a whole image at a time. To improve the localization of object boundaries, \citeauthor{chen2016deeplab}~\shortcite{chen2016deeplab} proposed a framework to combine Conditional Random Field (CRF) with FCNs. However, if we directly apply these deep-learning based segmentation algorithms to our problem of segmenting curve structures, it may produce non-curve segments because the CNNs are trained directly on the color/intensity images. In this paper, we will train CNNs on the curve-skeleton images to better learn the curve-geometry and curve-pattern features. More related to our work is Deep Skeleton~\cite{deepskeleton}, which also uses CNNs for skeleton extraction. However, Deep Skeleton is not specifically developed for curve structures and may produce many false positive skeletons. In the experiments, we include Deep Skeleton as a comparison method.
	
	\textbf{Curve-structure segmentation} from RGB or gray-scale images have been studied in many specific applications. For example, \citeauthor{lorigo2001curves}~\shortcite{lorigo2001curves} utilized an energy criterion based on intensity and local boundary smoothness to extract blood vessels in medical images. \citeauthor{tao2002using}~\shortcite{tao2002using} constructed a  statistical shape model to extract sulcal curves on the outer cortex of human brain. \citeauthor{zou2012cracktree}~\shortcite{zou2012cracktree} proposed a tree-based algorithm to detect curve-like cracks from pavement images. However, these methods all rely on specific assumptions in respective applications and it is not easy to extend the segmentation algorithm developed for one application to another application.
	
	Using computer vision and machine learning techniques to explore cultural heritage objects has been attracting more interest in recent years. However, most of them are focused on the classification and matching of  object fragments. For example, in \cite{smith2010classification,makridis2012automatic,rasheed2015archaeological}, various archaeological fragments are classified based on color and texture features.  In \cite{zhou2017identifying}, an extended Chamfer matching algorithms is developed to identify the design of a pottery sherd by matching the curve structures on the sherd to all the known designs, where the curve structures on the sherds are segmented with manual assistance. In this paper, we focus on accurate segmentation of curve structures on the surface of sherds, which is a fundamental step before the classification and matching.
	
	\section{Proposed Method}
    The proposed method consists of three steps. First, we train an FCN to detect the skeletons of the curve structures in the depth image. This FCN network also estimates a scale value at each detected skeleton pixel to reflect the curve width at this skeleton pixel. Second, we train a dense prediction convolutional network to identify and prune false positive skeleton pixels. Finally, we develop a scale-adaptive thresholding algorithm to recover the curve width and achieve the final segmentation of curve structures.
	
	\subsection{Step I: Detecting Curve Skeletons using FCN}
		In this paper skeletons are the center lines of the curve structures and they are of one-pixel width. By ignoring the curve width, the skeletons reflect the geometry and pattern of the curve structures. Therefore, in the first step, we train a FCN to detect the skeletons of the curve structures from an input depth image. Just like image segmentation, skeleton detection can be formulated as a pixel-labeling problem: skeleton pixel has a label $1$ and non-skeleton pixel has a label $0$.
		
\begin{figure}[htbp]
			\includegraphics[scale=1.1]{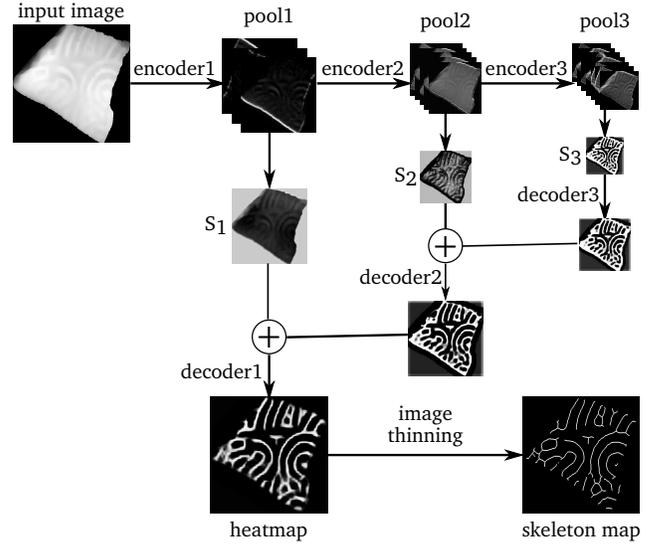}
			\caption{FCN used for skeleton detection.}\label{fig:step1}
		\end{figure}

	We design an FCN, as illustrated in Figure~\ref{fig:step1}, to label skeleton pixels. It follows the encoder-decoder architecture developed in~\cite{fcn}. Encoders 1 and 2 are small convnets made up of two $3\times 3$ convolutional layers, two ReLu layers and one $2\times 2$ max-pooling layer. Encoder 3 is a small convnet made up of three $3\times 3$ convolutional layers, three ReLu layers and one $2\times 2$ max-pooling layer. After an encoder, the image size will be reduced to 1/4. Therefore, the receptive field sizes of feature maps generated by the three encoders are $2\times 2$, $4\times 4$, and $8\times 8$, respectively. After each encoder, a fully connected layer is employed to match the number of feature maps with the number of labels. In order to generate pixelwise prediction result, the fully connected layers are implemented by $1\times 1$ convolutional layers. These results are denoted as $S_1$, $S_2$ and $S_3$, respectively, as shown in Figure~\ref{fig:step1}. Note that the size of $S_1$, $S_2$ and $S_3$ are successively downsampled by factors of 2, 4, and 8 from the original image size. The decoders are three deconvolution layers with a kernel size of $4\times 4$ and a stride of 2. The kernels are fixed to perform bilinear interpolation~\cite{xie2015holistically}. 
	
	The use of multiple encoders/decoders can extract image features in different levels of details. To make full use of all the extracted features, the decoders are organized in a way of stepwise accumulation when fusing them together.The output skeleton heat map $S$ can be computed by
	\begin{equation}\label{eq1}
		S =  softmax(\Psi^{(2)} (S_1 + \Psi^{(2)}(S_2 + \Psi^{(2)}(S_3)))  )
	\end{equation}
	where $\Psi$ indicates the upsampling operation performed by the decoders and its associated superscript is the upsampling factor, e.g.,  $\Psi^{(2)}$ indicates an upsampling of map by a factor of 2.  With the skeleton heat map $S$, we apply a common image thinning algorithm \cite{lam1992thinning} to generate the single-pixel width skeleton map.
	
	Inspired by~\cite{deepskeleton}, we can compare the three score maps $S_1$, $S_2$ and $S_3$ to estimate the scale at each detected skeleton pixel. The scale value at a skeleton pixel reflects the local curve width at this pixel. More specifically, since different encoders correspond to different receptive field sizes, at each pixel the receptive field size of the encoder with the largest score reflects the scale at this pixel. Before we compare the score of different maps, we need to first upsample them to the original image size. This way, the scale $s(x,y)$ at the skeleton pixel $(x,y)$ can be computed by
	\begin{equation}\label{eq2}
		s(x,y) = \arg\min_{k\in \{1, 2, 3\}} \hat{S}_k(x,y)
	\end{equation}
	where $\hat{S}_k=\Psi^{(2^{k})}(S_{k})$ is the upsampled score map of $S_k$. Later we will use the estimated scale values to help recover the curve width.
	
	\subsection{Step II: Refining Skeletons using Dense Prediction Convnet}
	
\begin{figure}[htbp]
		\centering
		\includegraphics[scale = 0.75]{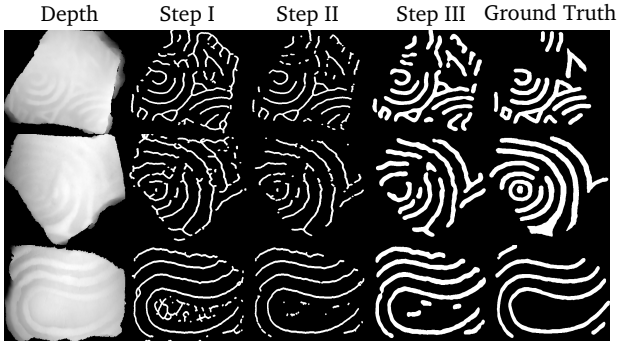}
		\caption{Example results after each step of the proposed method.}
		\label{step}
	\end{figure}

	While we expect the FCN trained in Step I can learn curve geometry and pattern features in detecting skeletons, we find that it still detects many false positive skeletons, as shown in Figure~\ref{step}. In this step, we further train a supervised classifier to identify and prune such false positives by learning more curve features. Specifically, for each skeleton pixel $(x,y)$ detected in Step I, we take a neighboring $45\times 45$ window in the original depth image around the pixel $(x,y)$ as the input and train a dense prediction convnet to determine whether $(x,y)$ is a true skeleton pixel or a false positive. 
	
	On real images, detecting a skeleton with small dislocation to its real position is totally fine and unavoidable -- even a manually labeled skeleton may not be perfectly aligned with the real center line of the curve structures. Therefore, our aim is not to directly train a hard classifier to distinguish skeleton pixels and non-skeleton pixels. Instead, we hope to train a soft classifier where a skeleton probability is outputted at each pixel.  To achieve this goal, in the training we transform a binary skeleton map to a skeleton probability map by
	\begin{equation}\label{eq3}
	D(x,y) = \frac{1}{1 + \underset{(x',y')\in P}{\min}\sqrt{(x-x')^2+(y-y')^2}}
	\end{equation}
	where $P$ is the set of skeleton pixels in the binary skeleton map. Using $D$ as output of the network, the binary classification problem is converted to a regression problem. Accordingly, we need to use a sigmoid function instead of softmax in the last layer of the proposed dense prediction convnet.
	
	In this paper, we propose to use a convnet consisting of three convolutional layers, three max-pooling layers and two fully connected layers. Its specific configuration is summarized in Table~\ref{convnet}. For a testing image, let the set of the skeleton pixels detected in Step I be $\hat{P}$ and the skeleton probability map generated by the prediction convnet in this step be $D$, we prune the low-probability ($<0.5$) skeleton pixels in $\hat{P}$ to achieve a refined set of skeleton pixels as
	\begin{equation}\label{eq4}
		P = \{(x,y)|(x,y)\in \hat{P}; ~~D(x,y)\geq 0.5\}
	\end{equation}
	Sample results of skeleton map after this step of refinement can be found in Figure~\ref{step}.
	
	\begin{table}
		\centering
		\caption{The configuration of network for Step II, where n, k, s, p stand for the number of outputs, kernel size, stride and padding size respectively.}\label{convnet}
		\begin{tabular}{|c | c|} 
			\hline
			\rule{0pt}{10pt} \textbf{Type} & \textbf{Configuration}\\
			\hline\hline
			\rule{0pt}{9pt} Sigmoid & -\\ \hline
			\rule{0pt}{9pt} Fully Connected & n:2\\ \hline
			\rule{0pt}{9pt} Dropout & ratio:0.5\\ \hline
			\rule{0pt}{9pt} Fully Connected & n:512\\ \hline
			\rule{0pt}{9pt} MaxPooling & k:$2\times 2$, s:2 \\\hline
			\rule{0pt}{9pt} Convolution & n:128, k:$3\times 3$, s:1, p:1 \\ \hline
			\rule{0pt}{9pt} Batch Normalization & -\\ \hline
			\rule{0pt}{9pt} MaxPooling & k:$2\times 2$, s:2 \\ \hline
			\rule{0pt}{9pt} Convolution & n:64, k:$3\times 3$, s:1, p:1 \\ \hline
			\rule{0pt}{9pt} Batch Normalization & -\\ \hline
			\rule{0pt}{9pt} MaxPooling & k:$2\times 2$, s:2 \\ \hline
			\rule{0pt}{9pt} Convolution & n:32, k:$3\times 3$, s:1, p:1 \\ \hline
			\rule{0pt}{9pt} Input & $45\times 45$ gray-scale image\\ \hline
		\end{tabular}
	\end{table}
	
	\subsection{Step III: Curve-Structure Segmentation by Recovering Curve Width}
	In this step, we recover the width of curve structures from the skeleton map derived in Step II, with the help of the scale values derived in Step I. Note that the width of the curve structures is not a constant and it may vary along the skeleton. Denote the original depth image by $I$ and let $P$ be the set of refined skeleton pixels detected on $I$ after Step II. For each skeleton pixel $(x,y)\in P$, we have a scale value $s(x,y)\in \{1,2,3\}$ derived in Step I. We construct the curve-structure segmentation, in the form of a binary map $C$ of the same size as $I$, using the following algorithm~\ref{alg:step3}.
	
	\begin{algorithm}
		\caption{Curve-Structure Segmentation by Recovering Curve Width}
		\textbf{Input:} Depth image $I$, Refined skeleton $P$, Scale values $s$
		\newline\textbf{Output:} \leftline{Binary segmentation map $C$}
		\vspace{-1em}
		\begin{algorithmic}[1]
			\STATE Initialize all the elements in $C$ to zero.
			\FOR {each skeleton pixel $(x,y) \in P$ }
			\STATE Compute neighborhood:\par
			$\mathbb{N} = \left \{ (x',y')|\sqrt{(x-x')^2+(y-y')^2} \leq 2^{s(x,y)}\right\}$.
			\FOR { each pixel $(x',y')\in \mathbb{N}$}
			\IF{$I(x',y')\geq\frac{I(x, y)+\min_{(x'',y'')\in \mathbb{N}} I(x'',y'')}{2}$}
			\STATE {$C(x',y')=1$}
			\ENDIF
			\ENDFOR
			\ENDFOR
		\end{algorithmic}
		\label{alg:step3}
	\end{algorithm}
	
		From the steps 3 and 5 of this algorithm, we can see that the curve width at each skeleton pixel is determined by both the scale value $s$ at this pixel and the depth values $I$ at and around this pixel. This algorithm does not require the detected skeleton to be exactly aligned with the center line of the curves -- a small dislocation of the skeletons may not change the final segmentation if the dislocated skeletons are still located inside the underlying curves. Sample results after Step III are shown in Figure~\ref{step}.
	
	\subsection{Design Matching}

\begin{figure}[htbp]
			\centering
			\includegraphics[scale=0.7]{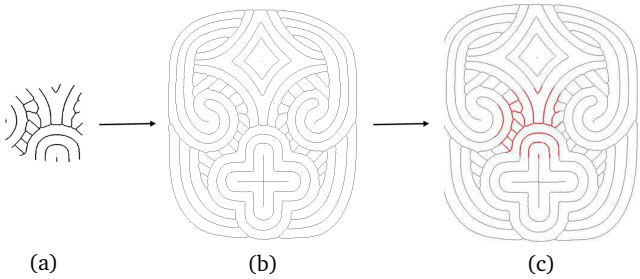}
			\caption{An illustration of design matching. (a) (Thinned) curve structures $U$ segmented on the sherd. (b) (Thinned)  full design $V$. (c)  Partial matching between $U$ to $V$ with minimal Chamfer distance. Original design illustration copyrighted by Frankie Snow. Used with	permission.}	\label{fig:Chamfer matching}
		\end{figure}

	One important application of the segmented curve structures in archeology is the task of design matching. In the later experiments, we will use this task to evaluate the performance of curve-structure segmentation. 
	As shown in Figure~\ref{fig:Chamfer matching}(c), a design is a full curve pattern of the paddle that are used for stamping the object surface. In the past decades, archaeologists have restored a small number of full designs by manually examining thousands of sherds~\cite{broyles1968reconstructed,snow1975swift}. The goal of design matching is to identify whether the segmented curve structures are originated from a known design. This is a classical partial matching problem and the key component is the definition of a matching score or distance. 
	
	In this paper, we use the classical Chamfer matching~\cite{barrow1977parametric,zhou2017identifying} for this purpose. As shown in Figure~\ref{fig:Chamfer matching}, we first thin both the segmented curve structures and the considered design into one-pixel wide skeletons and denote them as $U$ and $V$, respectively. We then transform $U$ to match the design $V$ and compute the Chamfer distance 
    \begin{equation}\label{eq5}
	d'_{CM}(U_{\bf T},V) = \frac{1}{|U|}\sum_{{\bf u} \in U_{\bf T}}\min_{{\bf v}\in V}\left \| {\bf u} - {\bf v} \right \|_2
	\end{equation}
	where $U_{\bf T}$ is the curve pattern $U$ after the transform $T$, ${\bf u} \in U_{\bf T}$ indicates all the skeleton-pixel coordinates ${\bf u} $ in the transformed partial pattern $U_{\bf T}$,
	and ${\bf v}\in V$ indicates all the skeleton-pixel coordinates ${\bf v}$ in the curve pattern $V$. $|U|$ is the total number of skeleton pixels in the partial pattern $U$. 
	Eq.~(\ref{eq5}) actually finds the nearest skeleton-pixel coordinate in $V$ for each skeleton-pixel coordinate in $U_{\bf T}$, records its Euclidean distance $ \left \| {\bf u} - {\bf v} \right \|_2$ and finally averages over all the skeleton-pixel coordinates in $U_{\bf T}$. 
	The matching distance between $U$ and $V$ is then defined by
	\begin{equation}\label{eq6}
		d(U,V) =\min_T d'_{CM}(U_{\bf T},V)
	\end{equation}
	with $T$ covers all possible translations and rotations. The scaling transforms is not considered here because both $U$ and $V$ have known actual sizes. 
	
	\section{Experiment}
	
	In this section, we validate the effectiveness of the proposed method from three perspectives. First, we evaluate the proposed method in terms of the classical metrics of precision, recall and F-measure and compare it
	against other six comparison methods. Second, we conduct experiment to justify the usefulness of each step in our method. Third, we evaluate the curve-structure segmentation results in the task of design matching.
	
	\subsection{Dataset}
	For this study, we collected the depth images of 1,174 pieces of pottery sherds that are excavated in various archaeological sites located in southeastern North America. We used a linear array 3D laser scanner, NextEngine, to get the point cloud of sherd surfaces with the resolution of 100 points per $mm^{2}$. Then their depth images are sampled with the same resolution, i.e., each pixel in depth image covers $0.01mm^{2}$. The average size of the collected depth images is $446\times421$. We have 530 of these depth images with manually labeled ground-truth curve-structure segmentations. Among all 530 images, we randomly pick 250 for training and the remaining 280 for testing.
	
	To train the FCN in Step I, we thin all the ground-truth curve structures to one-pixel width skeletons, using a standard image thinning algorithm~\cite{lam1992thinning}. Data augmentation is employed here to generate sufficient training data. Specifically, we first split the whole image into small blocks with a size of $100\times 100$. Then these blocks are rotated, scaled and flipped with the same scheme as in~\cite{deepskeleton}. Finally, 141,696 blocks are used in FCN training in Step I. As for the network training in Step II, we randomly take 44,906 window images with a size of $45\times 45$ around the skeleton pixels identified in Step I for training.
	
	\subsection{Implementation Details}
	
	For the purpose of better training, the parameters of encoders in the skeleton extraction network are initialized with the pre-trained FCN-8s model~\cite{fcn}. The parameters of decoder are fixed to perform bilinear interpolation~\cite{xie2015holistically}. The maximum number of training iterations is set as 20,000, with a mini-batch size of 10. The base learning rate is $1\times 10^{-7}$ and decays to $1\times 10^{-8}$ after 10,000 iterations. Momentum and weight decay are set to 0.9 and $5\times 10^{-4}$ respectively.
	
	Because the dense prediction convnet in Step II is relatively lightweight, we choose to train it from scratch. The maximum number of training iterations is set to 100,000, with a mini-batch size of 10. The base learning rate is $1\times 10^{-3}$, and it decays in an inverse way with the parameter $\gamma = 10^{-3}$ and $power = 0.75$. Momentum and weight decay are set to be the same as the FCN in Step I.
	
	\subsection{F-measure based Segmentation Performance}
	
	\begin{figure*}[htbp]
		\centering
		\includegraphics[width=\textwidth]{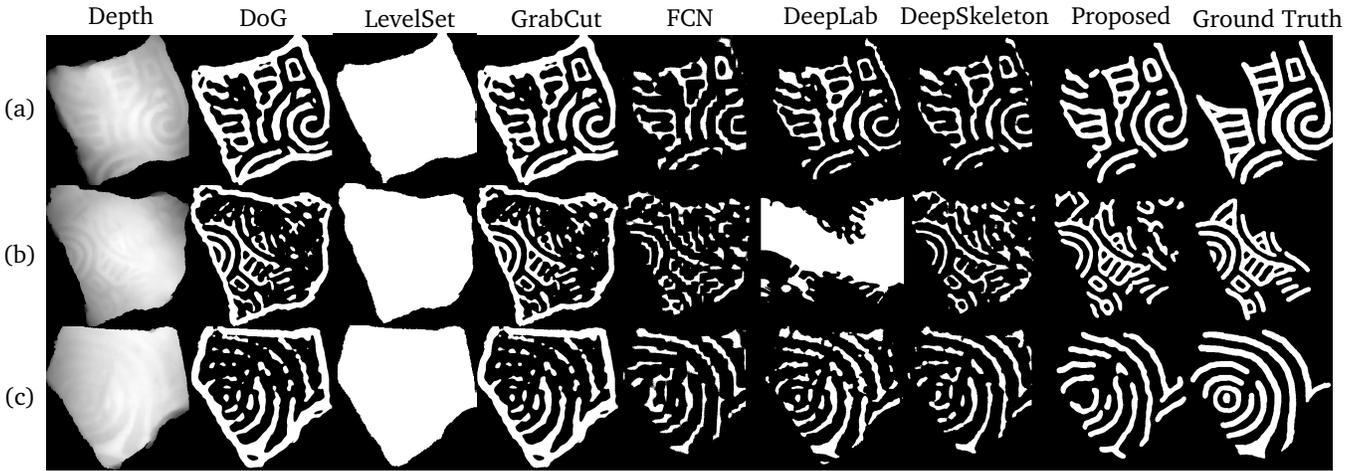}
		\caption{Examples of the curve-structure segmentation result from the proposed method and six comparison methods.}	\label{comparative}
	\end{figure*}

	To evaluate the effectiveness of our method of curve-structure segmentation, we select six widely-used segmentation methods for comparison -- Difference of Gaussian (DoG), Level Set~\cite{vese2002multiphase},  GrabCut~\cite{rother2004grabcut}, Fully Convolutional Network (FCN)~\cite{fcn}, Deep Skeleton~\cite{deepskeleton} and DeepLab~\cite{chen2016deeplab}. The experiment is conducted on the 280 testing images as described above, and the evaluation criteria is the traditional F-measure of $\frac{2 \cdot Precision \cdot Recall}{Precision+Recall}$.

	For most of these comparison methods, we keep the default settings in their source codes. But there are several exceptions need to be clarified. Since there is no default setting in DoG, we determine its parameters by trial-and-error. The best performing setting we found is: $k_1 = k_2 = 45$, $\sigma_1 = 11$, $\sigma_2 = 5$, where $k$ and $\sigma$ are the kernel size and standard deviation of Gaussian filters. The filtered images are transformed to curve maps with the threshold 1. In GrabCut, an initialization of the foreground object is required, for which we simply use the DoG result. In Deep Skeleton, we calculated the ground-truth scale maps by applying distance transform on ground-truth segmentation maps. Performance of all methods, averaged over all 280 testing images, are summarized in Table~\ref{evaluation}.
	
	\begin{table}
		\centering
		\caption{Precision, recall and F-measure of the proposed  method and six comparison methods, averaged over 280 test images.}	\label{evaluation}
		\begin{tabular}{cccc}
			\hline
			\rule{0pt}{10pt} \textbf{Methods}	&	\textbf{Precision}	&	\textbf{Recall}	& \textbf{F-measure} \\ \hline
			\rule{0pt}{9pt} DoG          & 0.366     & 0.774  & 0.490     \\
			\rule{0pt}{9pt} LevelSet     & 0.262     & 0.938  & 0.399     \\
			\rule{0pt}{9pt} GrabCut      & 0.357     & 0.671  & 0.448     \\
			\rule{0pt}{9pt} FCN          & 0.589     & 0.472  & 0.514     \\
			\rule{0pt}{9pt} DeepLab      & 0.585     & 0.670  & 0.583     \\
			\rule{0pt}{9pt} DeepSkeleton & 0.634     & 0.690  & 0.654     \\
			\rule{0pt}{10pt} \textbf{Proposed}     & 0.660 & 0.827 &\textbf{0.731}  \\ \hline
		\end{tabular}
	\end{table}

We can see that the proposed method achieves the best F-measure, and outperforms the second best (Deep Skeleton) by 7.7\%. Figure~\ref{comparative} shows the segmentation results on three sample images, using all seven methods. In these images, we can observe that DoG actually enhances the difference between adjacent pixels. As a purely low-level method, it may not capture deep and shallow curves simultaneously. GrabCut was initialized by DoG, but its performance becomes even worse. One major reason might be that the data and smoothness energy defined in GrabCut are not sufficiently discriminative to segment the curve structures and non-curve object surface in such a low-contrast image. This is probably the same reason that makes Level Set fail. As expected, the three CNN-based comparison methods, i.e., FCN, Deep Skeleton and DeepLab, normally achieve better performances than the low-level methods. However, their segmentation results usually contain many false positives and the boundaries of the segmented curve structures are quite rough. While the proposed method does not achieve the first place in either precision or recall, it achieves the best performance in final F-measure. 
	
	\subsection{Usefulness of Each Step}
	
	Intuitively, the three steps of our method can be replaced by other alternatives or simply ignored. To justify the usefulness of each step,  we design three additional experiments, in each of which, we modify or remove one step of the proposed method, and then check its influence to the final segmentation performance. 
	
\begin{figure}[htbp]
		\centering
		\includegraphics[width=0.48\textwidth]{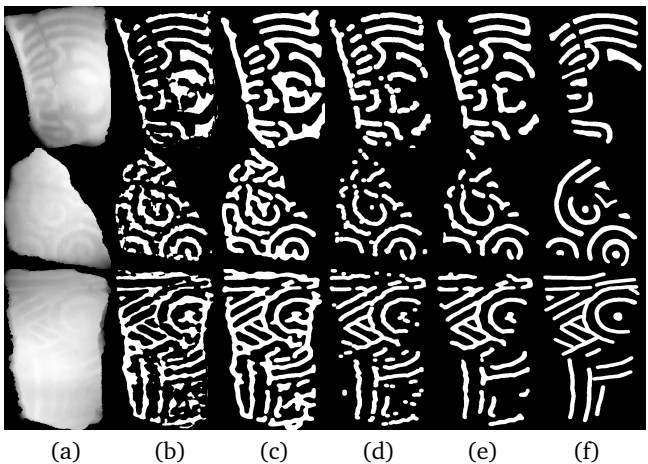}
		\caption{Sample segmentation results of the proposed method with modifications to each step. (a) Input depth image. (b) Segmentation result after modifying Step I. (c) Segmentation result after removing Step II. (d) Segmentation result after modifying Step III. (e) Segmentation result of the proposed method without any modification. (f) Ground-truth segmentation.} \label{ablation}
	\end{figure}

   \emph{Modifying Step I:}  Step I of the proposed method is skeleton extraction. Actually, the FCN we used in this step can be trained to produce curve-structure segmentation directly. However, we choose to extract skeletons first, and then take additional steps to recover the curve width. In this experiment, we make several adjustments in the FCN in Step I to let it output curve structures with width directly. For this purpose, we just use the ground-truth segmentation as the output for training and remove extra upsampling layers in FCN. All the implementation parameters keep unchanged. Sample results of this modified
	method are shown in Figure~\ref{ablation}(b) . We can see that these results contain more false positives and rougher segmentation boundaries. Quantitatively, F-measure of the proposed method decreases from 0.731 to 0.665 if we make this modification to Step I.

	\emph{Removing Step II:}  Step II of the proposed method employs a dense prediction convnet as a pixel-wise classifier to refine skeletons extracted by FCN in Step I. To justify its usefulness, we remove this step and recover curve width directly from the skeletons generated in Step I. Sample results are shown in Figure~\ref{ablation}(c). We can see that the removal of Step II leads to more false positives. Quantitatively, F-measure of the proposed method decreases from 0.731 to 0.662 if we remove Step II.
	
	\emph{Modifying Step III:}  Simple morphological dilation seems to be a very intuitive approach to recover curve width in Step III. In this experiment, we modify Step III by replacing it with a dilation operation with a radius of 15 pixels, which is the best parameter after we try and test all different values. Sample results are shown in Figure~\ref{ablation}(d). While the dilation produces very smooth curve structures, they do not align well with the ground truth. Quantitatively, F-measure of the proposed method decreases by 3.5\% if we make this modification to Step III.
	
	\subsection{Design-Matching Performance}
	
     In this experiment, we evaluate curve segmentation results in the task of design matching. We take the depth images of 292 sherds with known full designs and in total they come from 29 different designs.  
     The matching distance is the minimal Chamfer distance as defined above.
	
\begin{figure}[htbp]
	 		\centering
	 		\includegraphics[width=0.48\textwidth]{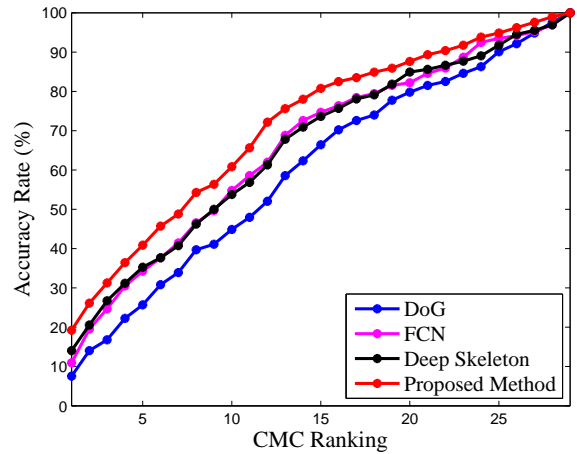}
	 		\caption{CMC curves of the proposed method and three comparison methods. }\label{CMC-curve}
	 	\end{figure}

	 We use the Cumulative Matching Characteristics (CMC) ranking metric to evaluate the design-matching performance. For each sherd curve-pattern $U$, we match it against all 29 designs by Chamfer matching. We then sort these 29 designs in terms of the matching distance and pick the top $L$ matching designs with the smallest matching distances. If the ground-truth design of this sherd is among the identified top $L$ designs, we count it as a correct matching under rank $L$.  We repeat this for all $292$ sherds and calculate the accuracy, i.e., the percentage of the correctly matched sherds, under each rank $L$, $L=1, 2,\cdots, 29$. This way, we can draw a CMC curve in terms of rank $L$ as shown in Figure~\ref{CMC-curve}, which reflects the performance of curve-structure segmentation -- The higher the CMC curve, the better the segmentation performance. 
	 
	Besides the proposed method, we select three other representative comparison segmentation methods for performance evaluation in this experiment. These three comparison methods are DoG, FCN and Deep Skeleton. Figure~\ref{CMC-curve} shows the CMC curves of the proposed method and these three comparison methods in the task of design matching. The proposed method achieves a CMC rank-1 rate of 20\% and a CMC rank-15 rate of 78\%, which are much better than the other three comparison methods.

	\section{Conclusion}
	In this paper, we put forward a novel and challenging image segmentation problem: weak curve-structure segmentation from noisy depth images, which has important applications in archeology for exploring large collections of fragmented cultural heritage objects. We developed a new three-step supervised-learning based method to address this problem, by first extracting and refining the skeletons of underlying curve structures and then producing the final segmentation by recovering the curve width at each skeleton pixel. In the experiment, we tested the proposed method on a dataset of depth images scanned from unearthed pottery sherds from southeastern North America. We found that the proposed method performs better than several widely used low-level and deep-learning based image segmentation methods in terms of F-measure.
	
	\textbf{Acknowledgment} This work was partly supported by NSF-1658987 and NSFC-61672376.

	\bibliographystyle{aaai}
	\bibliography{Lu}
\end{document}